# #phramacovigilance - Exploring Deep Learning Techniques for Identifying Mentions of Medication Intake from Twitter

**Debanjan Mahata**
Bloomberg
New York, U.S.A
dmahata@bloomberg.net

**Jasper Friedrichs**
Mountainview,
C.A., U.S.A
jasper.friedrichs@gmail.com

**Rajiv Ratn Shah**
IIIT-Delhi,
New Delhi
rajivratn@iiitd.ac.in

**Hitkul**
IIIT-Delhi,
New Delhi
hitkuljangid@gmail.com

## Abstract

Mining social media messages for health and drug related information has received significant interest in pharmacovigilance research. Social media sites (e.g., Twitter), have been used for monitoring drug abuse, adverse reactions of drug usage and analyzing expression of sentiments related to drugs. Most of these studies are based on aggregated results from a large population rather than specific sets of individuals. In order to conduct studies at an individual level or specific cohorts, identifying posts mentioning intake of medicine by the user is necessary. Towards this objective, we train different deep neural network classification models on a publicly available annotated dataset and study their performances on identifying mentions of personal intake of medicine in tweets. We also design and train a new architecture of a stacked ensemble of shallow convolutional neural network (CNN) ensembles. We use random search for tuning the hyperparameters of the models and share the details of the values taken by the hyperparameters for the best learnt model in different deep neural network architectures. Our system produces state-of-the-art results, with a micro-averaged F-score of 0.693.

## 1 Introduction

Social media is a ubiquitous source of information for a variety of topics. These platforms have shown unprecedented growth in terms of users. Information related to daily events and personal rants, as well as expressions related to the intake of medicine and adverse drug reactions are readily available in publicly accessible social media channels such as Twitter[1], DailyStrength[2] and MedHelp[3], among others. Huge amounts of data made available on these platforms have become a useful resource for conducting public health monitoring and surveillance, commonly known as pharmacovigilance (Sarker et al., 2015).

As of 1 January 2018, Twitter has 330 million monthly active users[4]. A study conducted in 2011 found a total of 620 breast cancer groups on Facebook that had 1,090,397 members. These groups aimed at spreading awareness, raising funds and providing support to the sufferers[5]. Moreover, a study showed that 42% of individuals viewing health information on social media look at health-related consumer reviews and 32% of United States users post about their friends' and family's health experiences on social media[6]. Doctors and hospitals also use social media to better serve their patients. Therefore, social media can have a big impact on people's medicine and health choices.

Given the abundance of data and adoption of the social media platforms for sharing health related information, attempts have been made to mine content from posts mentioning medications in order to identify adverse drug reactions (Nikfarjam et al., 2015), abuse (Hanson et al., 2013), and user sentiment (Korkontzelos et al., 2016). However, all these studies are based on aggregated results from large set of content that mentions a medicine/drug, without taking into account whether the user has actually consumed it. Without this knowledge, a true assessment of the effects of medication intake, in general, and how it affects a specific group of users cannot be done. In order to leverage social media data for per-forming such assessments and studying targeted groups, it is necessary to develop systems that can automatically distinguish posts that expresses personal intake of medicine from those that do not. This would further facilitate high precision pharmacovigilance.

---

[1] http://twitter.com
[2] https://www.dailystrength.org/
[3] http://www.medhelp.org/
[4] https://www.omnicoreagency.com/twitter-statistics/ Accessed on 23 March 2018
[5] http://www.adweek.com/digital/pius-boachie-guest-post-3-ways-social-media-revolutionized-medical-care/ Accessed on 23 March 2018
[6] https://getreferralmd.com/2017/01/30-facts-statistics-on-social-media-and-healthcare/ Accessed on 23 March 2018

***Objective*** - In this work, we concentrate on Twitter as the social media channel for our experiments. The key to the process of identifying tweets mentioning personal intake of medicine and to draw insights from them is to build accurate text classification systems. The main objective of the task presented in this paper is to categorize short colloquial tweets into one of the three classes:

- *personal medication intake (Class 1)* - tweets in which the user clearly expresses a personal medication intake/consumption (e.g., *I had the worst headache ever and I just took an AdvilRelief #advil and now I feel so much better thank*).

- *possible medication intake (Class 2)* - tweets that are ambiguous but suggest that the user may have taken the medication (e.g., *I should have taken advil on friday then i might have actully had an amazing weekend.. instead of throwing up 20 times a day #advil, not this time*).

- *non-intake (Class 3)* - tweets that mention medication names but do not indicate personal intake (e.g., *Understand the causes and managing #Migraine Madness #aspirin #diet #botox #advil #relpax #headache*).

***Challenges*** - Mining social media posts comes with unique challenges. Microblogging websites like Twitter pose challenges for automated information mining tools and techniques due to their brevity, noisiness, idiosyncratic language, unusual structure and ambiguous representation of discourse. Information extraction tasks using state-of-the-art natural language processing techniques, often give poor results for tweets. Abundance of link farms, unwanted promotional posts, and nepotistic relationships between content creates additional challenges (Mahata et al., 2015). In addition to this, for an automated system, there is an implicit challenge associated with distinguishing between *Class 1* and *Class 2* of the task. This is primarily because it is often difficult to understand from the informally expressed tweets whether a user has actually consumed a medicine or has just mentioned it.

***Contributions*** - Some of the contributions that we make in this paper are:

- We explore the performances of different standard deep neural network architectures on the given task and share our learnings.

- Design and train a *stacked ensemble of shallow CNN ensembles*, which gives state-of-the-art performance on the presented task.

- We make all the trained models publicly available and share their corresponding best performing hyperparameters[7].

## 2 Background and Related Work

In this section, we present the existing body of scientific work closely related to ours, along with a brief background of the deep learning architectures that we use for our experiments. We share the rationale behind the choice of the models and explain how our approach of training a *stacked ensemble of CNN ensembles* is different from existing deep learning architectures.

***Existing Approaches and Deep Learning*** - The problem of detecting personal intake of medicine from Twitter was first introduced at *SMM4H* workshop (Sarker and Gonzalez-Hernandez) as a shared task. The workshop introduced different approaches and presented current state-of-the-art performances on the given dataset and task. With the exception of *NRC-Canada* (Kiritchenko et al.), that implements a Support Vector Machine classifier using a variety of surface-form, sentiment, and domain-specific features, all the other systems attempt to solve the task using convolutional neural networks (CNNs). *UKNLP* (Sifei et al.), trained a CNN model with attention mechanism. *CSaRUS-CNN* (Arjun et al.), uses a cost sensitive and random undersampling variants of CNNs. *TurkuNLP* (Kai et al.), developed an ensemble of neural networks with features generated by word and character-level convolutional neural network channels, and a condensed weighted bag-of-words representation. There is a clear indication of ensembles and CNNs being the dominant strategy for the top teams.

---

[7]link not provided in order to maintain the anonymity of the publication

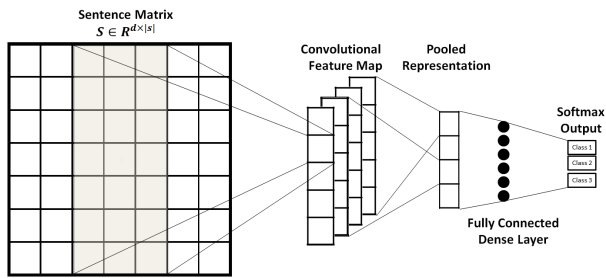

Figure 1: Shallow convolutional neural network (CNN) architecture with a single input channel.

***Convolutional Neural Network*** - A convolutional neural network (CNN) is a deep learning architecture, that has shown strong performance on text classification tasks. The basic building blocks of a standard CNN architecture are - *convolutional layers*, *pooling layers* and *fully connected layers*. The *convolutional layer* consists of filters and feature maps. The pooling layer follow a sequence of one or more convolutional layers and aid in consolidating the features learned and expressed in the previous layer's feature map. The final layer is a fully connected, flat feed-forward neural network used for producing non-linear combination of features and for making predictions by the network. This layer may have a non-linear activation function or a softmax activation in order to output probabilities of class predictions. For our CNN model, we use a softmax activation in the final layer in order to get the probabilities of the three classes for a given input. In order to get a more lucid explanation of CNNs, please refer to (Karpathy, 2016). The architecture of a basic shallow CNN for the classification task, as used in our experiment is very similar to (Severyn and Moschitti, 2015; Kim, 2014) and is shown in Figure 1.

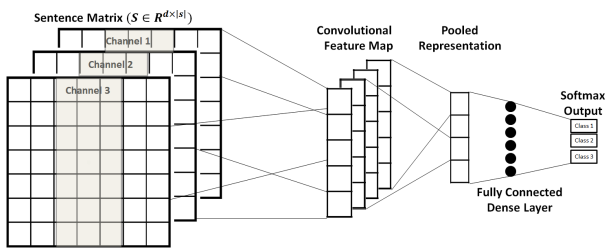

Figure 2: Shallow convolutional neural network architecture with three input channels.

Convolutional neural networks are effective in text classification tasks primarily because they are able to pick out salient features (e.g., tokens or sequences of tokens) in a way that is invariant to their position within the input sequence of words. In this paper we explore two different CNN architectures. In addition to the standard shallow CNN we also explore a multi-channel CNN as described in (Kim, 2014). A multi-channel convolutional neural network for document classification uses multiple versions of the standard CNN model with different sized filters. This allows the document to be processed at different resolutions or different n-grams (groups of words) at a time, whilst the model learns how to best integrate these interpretations[8]. Figure 2 shows the architecture of the multi channel CNN that we use in our experiment.

Recent studies such as (Kim, 2014) explored shallow single layer convolutional neural network as well as multi-channel CNN for sentence classification. Lin *et al.* (Lin et al., 2014) used deep neural network techniques that includes a CNN to detect user-level psychological stress from social media. Le *et al.* (Le et al., 2017) studied standard text classification tasks using CNN models and showed that deep CNNs perform well on all the tasks when the text input is treated as a sequence of characters. Contrary to it they also showed that going deeper with respect to convolutions does not always lead to the best solution, and a shallow but wider CNN can outperform the deeper ones when the text input is treated as a sequence of words. Szegedy *et al.* (Szegedy et al., 2015) found that their shallow word models outperform deeper models.

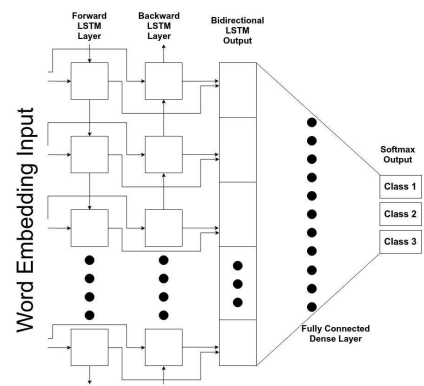

Figure 3: Bi-directional LSTM RNN architecture.

***Bi-directional LSTM Recurrent Neural Network*** - A Recurrent Neural Network (RNN) (Elman, 1990) is a neural network architecture that can process sequences of arbitrary length. In RNNs, a state at time $t$ is a function of input value at time $t$ and state at time $t − 1$. Due to this ability, RNN has given state-of-the-art performance in sequential learning tasks, especially in sentence classification (Zhou et al., 2016), speech recognition (Schuster, 1999), protein structure prediction (Baldi et al., 1999) and phoneme classification (Graves and Schmidhuber, 2005). However, on using a conventional nonlinear transition function, RNN suffers from the problem of exploding or vanishing gradients (Hochreiter et al., 2001) i.e., components of gradient vector starts growing or decaying exponentially. To solve this problem, Long Short-Term Memory (LSTM) units were introduced in 1997 (Hochreiter and Schmidhuber, 1997). LSTM is an adaptive gating mechanism. It decides what degree of previous state data and input data should be memorized for creating the new state.

---

[8]https://machinelearningmastery.com/develop-n-gram-multichannel-convolutional-neural-network-sentiment-analysis/

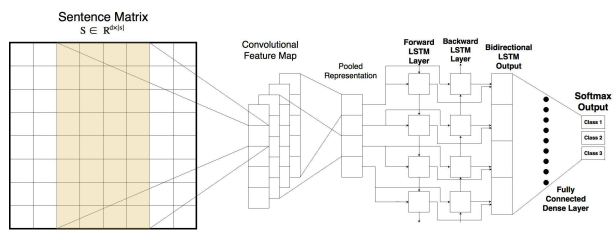

Figure 4: CNN+Bi-directional LSTM RNN architecture.

Knowledge of future along with past can be helpful for sequence modeling tasks. This is only possible for tasks in which all the time steps of an input sequence is available. Words and sentences which do not make sense standalone, become meaningful in presence of future context (Graves and Schmidhuber, 2005). Based on this concept, Bidirectional LSTMs (BLSTM) were proposed (Schuster and Paliwal, 1997). BLSTM is an extension of LSTM in which two LSTM models are trained on the input sequence. The first on the input sequence as-is and the second on its reversed copy. This can provide additional context to the network and result in faster and sometimes better learning. At any time t, a state is the function of forward and backward states. This allows BLTSM RNNs to learn from both past and future context. These properties make BLSTM suitable for sentence classification tasks (Liu et al., 2016b; Ding et al., 2018; Tai et al., 2015).

In the SMM4H shared task 2 (Sarker and Gonzalez-Hernandez), none of the teams successfully explored BLTSMs and achieved high performance. In our experiments we train a BLSTM model with the architecture as shown in Figure 3. Additionally, we also combine the power of CNNs and BLSTMs by training a hybrid architecture as shown in Figure 4.

*Deep Learning Ensembles* - Often, one solution to a complex problem does not fit to all scenarios. Thus, researchers use ensemble techniques to address such problems. Historically, ensemble learning has proved to be very effective in most of the machine learning tasks including the famous winning solution of the Netflix Prize (Wu, 2007). Ensemble models can offer diversity over model architectures, training data splits or random initialization of the same model or model architectures. Multiple average or low performing learners are combined to produce a robust and high-performing learning model. Some of the popular ensemble approaches are *bagging* and *boosting* (Quinlan et al., 1996). *Stacked ensembles* are also one of the widely used ensemble technique that we leverage in this work. Stacked ensembles finds the optimal combination of a collection of prediction algorithms using a process called stacking (Van der Laan et al., 2007). It involves training a second-level "metalearner" to find the optimal combination of the first level base learners. Unlike bagging and boosting, the goal in stacking is to put together strong, diverse sets of learners.

Zhou *et al.* (Zhou et al., 2002) presented a neural network ensemble and proposed, that many neural networks can be jointly used to solve a problem efficiently. For instance, Deng and Platt (Deng and Platt, 2014) use an ensemble of deep learning models for speech recognition. Wang *et al.* (Wang et al., 2008) presented *ensemble of classifier* approaches for biomedical named entity recognition by combining *generalized winnow*, *conditional random fields*, *support vector machine*, and *maximum entropy* through three different strategies - arbitration rules, stacked generalization (class-stacking and class-attribute-stacking) (Van der Laan et al., 2007), and cascade generalization (Gama and Brazdil, 2000). Ensemble techniques in general have shown to perform well in biomedical entity extraction (Ekbal and Saha, 2013) and named entity recognition (Sikdar et al., 2012; Speck and Ngomo, 2014). Furthermore, stacked ensemble techniques were very useful in different healthcare applications (Dinakar et al., 2014). A recent work (Liu et al.,

2016a), proposed several ensemble approaches including stacked generalization that effectively distinguishes between adverse drug events (ADEs) and non-ADEs from informal text in social media. Our literature review confirms that leveraging social media data using ensemble and neural network techniques is very beneficial in healthcare applications.

In order to take the best of all worlds, we combine the power of CNNs and ensemble techniques to train ensembles of shallow CNNs with different hyperparameters that are randomly chosen. In order to leverage the effectiveness of ensembles to a great extent, we also come up with the design of a new architecture of *stacked ensemble of CNN ensembles*. This architecture combines the best CNN ensembles that are trained using random parameter choices, and puts them together in a single framework for performing the final predictions. We obtain state-of-the-art results using the ensemble approaches. Deep learning models have been rarely combined using a stacked ensemble architecture. To our knowledge, this is the first work in this domain that uses a stacked ensemble for combining the predictions of several deep learning models trained on different hyperparameter settings that are randomly chosen. Also, we use an ensemble of several shallow CNNs as the first level learners and create an ensemble of the top performing CNN ensembles in the second level as opposed to using metalearners. We believe our architecture is a contribution to the field as it is a modified form of *stacked generalization of ensembles*, and shows best performance for the presented task. Next we explain the architecture of a *stacked ensemble of shallow CNN ensembles*.

## 3 Stacked Ensemble of Shallow Convolutional Neural Networks

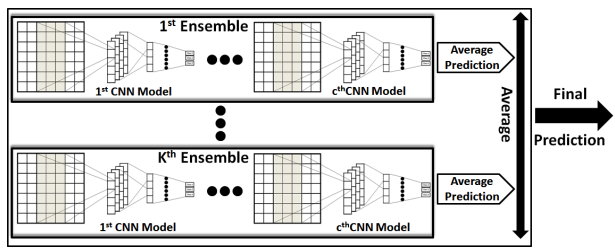

Figure 5: A stacked ensemble of ($K \times c$) shallow convolutional neural networks.

A stacked ensemble of shallow CNNs is a large ensemble classifier comprising of smaller ensembles stacked together, prioritized by their performance, with the underlying classifier being a standard shallow CNN model. In order to train such an ensemble model we enlist the generic steps in Algorithm 1.

**Algorithm 1:** Steps for training a *stacked ensemble of shallow CNN ensembles*.

1. Permute hyperparameter options to be considered to a set $H$ of hyperparemter configurations.
2. Randomly select and remove a hyperparameter configuration from $H$ and train a shallow CNN model on each fold while performing $c$-fold cross validation on the training dataset.
3. The output prediction of each model trained on each fold is averaged to get the final prediction of an ensemble of $c$ CNN models, $prediction_i^{ensemble} = \frac{1}{c} \sum_{j=1}^{c} prediction_{i_j}$.
4. Repeat step 2 and 3 $n$ times.
5. Sort the $n$ ensembles in terms of their performance on the metric suitable for the classification task.
6. Choose top $k$ ensembles based on their performance on the training dataset to form the final stacked ensemble of $k$ CNN ensemble models.
7. The final output prediction is given by the average of the predictions made by each of the top $k$ ensembles, $prediction_k^{stacked} = \frac{1}{k} \sum_{l=1}^{k} prediction_{top_l}^{ensemble}$.

Figure 5, shows a high level architecture of *stacked ensemble of shallow CNN ensembles* that we use in predicting the outcome of the task. Next, we share the detailed settings, output and analysis of our experiments.

## 4 Experiments

In this section, we present the details of the experiments that we perform. We explore different standard deep learning architectures for CNNs and RNNs as explained in Section 2 and show that the best results are obtained using ensemble strategies.

One of the key component of the input fed to the deep learning models are pre-trained word embedding vectors that are used for representing each word of the input tweets by a dense real valued vector. Since the dataset on which we train our model is relatively small we use the pre-trained word embeddings in order to prevent overfitting. This practice is commonly known as *transfer learning*[9]. Next, we give a brief overview of the different word embedding models that we use.

### 4.1 Word Embedding Models

*Godin Twitter Word2Vec Embedding* - This is a popular Word2Vec model trained on 400 million

---
[9] ftp://ftp.cs.wisc.edu/machine-learning/shavlik-group/torrey.handbook09.pdf

tweets, in which the word vectors are inferred using skip-gram architecture and negative sampling, along with the default parameters of the word2vec tool (Godin et al., 2015). The trained model has a dimension of 400 with a vocabulary of 3,039,345 unique words and is publicly available[10].

***Shin Twitter Word2Vec Embedding*** - These word embeddings are trained on tweets collected by the Archive Team[11], and has a vocabulary of 3,676,786 unique words. The word embeddings are trained by the original implementation of Word2Vec[12] from Google using skip-gram and negative sampling, with the default parameter settings. Four sets of embeddings with different dimensions (50, 100, 200, 400) are trained. These models are not publicly available and were acquired by directly contacting the authors of the paper (Shin et al., 2016).

***Twitter Glove Embedding*** - Glove is a popular algorithm for generating word embedding vectors (Pennington et al., 2014). The authors of Glove has made publicly available, several word embeddings trained on different types of datasets in their website[13]. Among these embeddings we use the one trained on 2 billion tweets with a vocabulary of 1.2 million unique words. The models are available with different dimensions (25, 50, 100, 200). For our experimental purpose we use the one with a dimension of 200.

***Drug Chatter Word2Vec Embedding*** - This is an embedding model trained by us on a dataset primarily containing drug-related chatter from Twitter, that is publicly shared[14] by the authors of (Sarker and Gonzalez, 2017). The original dataset consists of 267,215 Twitter posts made during the four-month period of November, 2014 to February, 2015, and mentions over 250 drug-related key words. However, we were unable to collect all the tweets as a huge chunk of them were not accessible anymore using the Twitter API[15]. We were only able to collect 155,394 tweets. We used the Word2Vec module provided with Gensim[16] for training a word2vec model on the collected tweets, using the skip-gram architecture and negative sampling. We didn't tune any hyperparameters and used the default settings as provided by the module. The vocabulary of the model is 47,943 unique words.

The dataset used for training the models is explained next.

### 4.2 Dataset

The dataset used in this paper is publicly available and can be obtained from the 2nd Social Media Mining for Health Applications Shared Task at AMIA 2017 website[17]. The organizers of the task provided 8,000 annotated tweets as a training dataset and 2,260 additional tweets as development dataset. We collected the tweets using the script provided along with the dataset, by querying Twitter's API. However, we could not collect all the tweets as some of them were not available at the time when we executed our collection process. The organizers also shared the test data. The test dataset consisted 7,513 tweets. A distribution of tweets provided for each class is shown in Table 1.

|  | Class 1 | Class 2 | Class 3 | Total |
|---|---|---|---|---|
| **Train** | 1,847 | 3,027 | 4,789 | 9,663 |
| **Test** | 1,731 | 2,697 | 3,085 | 7,513 |

Table 1: Dataset distribution. Classes 1, 2 and 3 represents *personal medication intake*, *possible medication intake* and *no medication intake*, respectively.

We train several deep learning models by combining the training and development data provided and treating it as our training dataset consisting of 9663 tweets. We learn our models using 5-fold cross validation. We explain the training process and the corresponding evaluation results obtained on the test data for each model in Section 4.4.

### 4.3 Evaluation Metrics

The evaluation metric used was micro-averaged F-score ($\mathcal{F}_{1+2}$) of the Class 1 (personal medication intake) and Class 2 (possible medication intake), for assessing the performance of our model, as used in the *Social Media Mining for Health* shared task (Sarker and Gonzalez-Hernandez). The equation for calculating the micro-averaged F-score for classes 1 and 2, which in turn depends on micro-averaged precision ($\mathcal{P}_{1+2}$) and recall ($\mathcal{R}_{1+2}$) for classes 1 and 2, is shown in equations 1, 2 and 3,

---
[10]https://www.fredericgodin.com/software/
[11]archive.org/details/twitterstream
[12]code.google.com/p/word2vec
[13]https://nlp.stanford.edu/projects/glove/
[14]http://diego.asu.edu/Publications/Drugchatter.html
[15]https://developer.twitter.com/en/docs
[16]https://radimrehurek.com/gensim/models/word2vec.html
[17]https://healthlanguageprocessing.org/sharedtask2/

respectively.

$$\mathcal{F}_{1+2} = \frac{2 * Precision_{1+2} * Recall_{1+2}}{Precision_{1+2} + Recall_{1+2}} \quad (1)$$

$$\mathcal{P}_{1+2} = \frac{TP_1 + TP_2}{TP_1 + FP_1 + TP_2 + FP_2} \quad (2)$$

$$\mathcal{R}_{1+2} = \frac{TP_1 + TP_2}{TP_1 + FN_1 + TP_2 + FN_2} \quad (3)$$

where, $TP_i$ is the number of True Positives for $Class\ i$; $TN_i$ is the number of True Negatives for $Class\ i$; $FP_i$ is the number of False Positives for $Class\ i$; $FN_i$ is the number of False Negatives for $Class\ i$.

### 4.4 Training Deep Learning Models

| Hyperparameter | Range |
|---|---|
| Word embedding choices for Stacked Ensemble of Shallow CNN Ensembles | Godin, Shin |
| Word embedding choices for other models | Godin, Shin, Glove, Drug |
| No. of Filters | 100, 200, 300, 400 |
| Filter Sizes for Stacked Ensemble of Shallow CNN Ensembles on each fold of 5-folds | [1,2,3,4,5], [2,3,4,5,6], [3,4,5,6,7], [1,2,2,2,3], [2,3,3,3,4], [3,4,4,4,5], [4,5,5,5,6] |
| Filter sizes for the CNN models | [1, 2, 3, 4, 5, 6] |
| Dense Layer Size | 100, 200, 300, 400 |
| Dropout Probability | 0.4, 0.5, 0.6, 0.7, 0.8, 0.9 |
| Batch sizes for Stacked Ensemble of Shallow CNN Ensembles | 50, 100, 150 |
| Batch sizes for other models | 8, 16, 32 |
| Number of BLSTM Units | [8, 16, 32, 64, 128, 256, 512] |
| Learning Rate | 0.0001, 0.001 |
| Adam beta2 | 0.9, 0.999 |

Table 2: Hyperparameter ranges used for random search permutations.

In order to get the best results from any classification model, hyperparameter tuning is a key step and deep neural networks are no exception. While the existing literature offers guidance on practical design decisions, identifying the best hyperparameters of a deep neural network requires experimentation. This requires evaluating trained models on a cross-validation dataset and manually choosing the hyperparameters that produce the best results. Automated hyperparameter searching methods like *grid search*, *random search*, and *bayesian optimization* methods are also commonly used. For our experiment we use random search (Bergstra and Bengio, 2012), to explore the hyperparameters of the different architectures. The full list of hyperparameters provided to the random search process is shown in Table 2.

We use NLTK[18] and it's tweet tokenizer for all our data preprocessing and cleaning activities. We do not remove stopwords. Each document in our training and test dataset is converted to a fixed size document of 47 words/tokens. Each word in the input tweet is represented by its corresponding embedding vector, when present in the vocabulary of the word embedding model that is randomly chosen as a hyperparameter. Tweets are mapped to embedding vectors and are available as a matrix input to the model. The *stacked ensemble of shallow CNN ensembles* is trained using Tensorflow[19] and all the other models are trained using Keras[20] with a Tensorflow backend.

***Stacked Ensemble of Shallow CNN Ensembles*** - We train a standard shallow CNN model, on each fold while performing 5-fold ($c = 5$) cross validation on our training dataset. We take the output prediction of each of these models trained on each fold and average them to create an ensemble of 5 models. We generate a set of 16,128 hyperparameter configurations (from Table 2) and randomly select a subset of 925 configurations to train such ensembles. For the final prediction we sort the ensembles in order of their decreasing performance on the training dataset and take the top $k$ ensembles. We take the prediction of each of the $k$ ensembles and average them to get the final prediction from our stacked ensemble of shallow CNNs (see Table 3 for the choices of $k$). In general, we can take top $k$ such ensembles and create a stacked ensemble of top $k$ ensemble of shallow CNNs. Table 3, shows the detailed performance results on the test data for the different top $k$ values for *stacked ensemble of shallow CNN ensembles*.

| Method | ($\mathcal{P}_{1+2}$) | ($\mathcal{R}_{1+2}$) | ($\mathcal{F}_{1+2}$) |
|---|---|---|---|
| CNN | 0.670 | 0.650 | 0.658 |
| Multichannel CNN | 0.650 | 0.665 | 0.656 |
| BLSTM | 0.675 | 0.690 | 0.679 |
| CNN + BLSTM | 0.675 | 0.700 | 0.687 |

Table 4: Performances of CNN, multi channel CNN, BLSTM and CNN+BLSTM on the test data.

***Other Models*** - All the remaining architectures are trained using 5-fold cross validation, with a set of hyperparameters randomly chosen from the provided choices as given in Table 2. We train 250 models for each architecture and choose the model giving the best performance for evaluation against the test data. While training a multi channel CNN, we make sure that the dimensions of the word embeddings chosen in each channel are the same. Therefore, we do not use the Glove embeddings as they are of 200 dimensions, while the other embeddings have dimensions of 400. Tables 4 shows the performances of the the best chosen model for each architecture on the test data.

---
[18]https://www.nltk.org/
[19]https://www.tensorflow.org/
[20]https://keras.io/

|  | Recall | | | Precision | | | F1 | | | $\mathcal{R}_{1+2}$ | $\mathcal{P}_{1+2}$ | $\mathcal{F}_{1+2}$ |
|---|---|---|---|---|---|---|---|---|---|---|---|---|
|  | 1 | 2 | 3 | 1 | 2 | 3 | 1 | 2 | 3 | | | |
| *top 3* | 0.699 | 0.640 | 0.842 | 0.701 | 0.732 | 0.758 | 0.700 | 0.682 | 0.798 | 0.718 | 0.663 | 0.690 |
| *top 10* | 0.700 | 0.644 | 0.848 | 0.708 | 0.730 | 0.764 | 0.704 | 0.684 | 0.804 | 0.721 | 0.666 | 0.692∗ |
| *top 20* | 0.697 | 0.636 | 0.851 | 0.706 | 0.731 | 0.759 | 0.702 | 0.680 | 0.802 | 0.720 | 0.660 | 0.689 |

Table 3: Evaluation of stacked ensembles of shallow CNN ensembles on test data. ∗ marks the state-of-the-art micro averaged F1 on the task's dataset achieved by our best model.

| Method | Hyperparameters of the Best Performing Model |
|---|---|
| **CNN** | Epochs: 7, Batch Size: 16, Filter Size: 2, No. of Filters:100, No. of Dense Outputs: 100, Word Embedding: Godin, Dropout Probability: 0.7 |
| **Multi-channel CNN** | Epochs: 5, Batch Size: 32, Filter Size Channel 1: 2, Filter Size Channel 2: 3, Filter Size Channel 3: 1, No. of Filters Channel 1:100, No. of Filters Channel 2: 400, No. of Filters Channel 3: 200, Word Embedding Channel 1: Drug, Word Embedding Channel 2: Godin, Word Embedding Channel 3: Godin, Dropout Probability Channel 1: 0.9, Dropout Probability Channel 2: 0.7, Dropout Probability Channel 3: 0.8 |
| **BLSTM** | Epochs: 9, Batch Size: 8, No. of BLSTM Units: 128, Word Embedding: Godin, Dropout Probability: 0.7 |
| **CNN+BLSTM** | Epochs: 12, Batch Size: 8, No. of BLSTM Units: 16, CNN Filter Size:4 , CNN No. of Filters: 400, Word Embedding: Godin, Dropout Probability BLSTM: 0.2, Dropout Probability CNN: 0.7 |

Table 5: List of hyperparameters for the best performing model of each deep learning method.

| id | Word Embedd. Model | Batch Size | Filter Sizes | Drop. Prob. | Dense Layer Size | No. of Filters | Learning Rate | Adam beta2 | $\mathcal{F}_{1+2}$ (train) | $\mathcal{F}_{1+2}$ (test) |
|---|---|---|---|---|---|---|---|---|---|---|
| 1932 | godin | 50 | [3, 4, 4, 4, 5] | 0.4 | 100 | 400 | 0.0001 | 0.9 | 0.7281 | 0.6886 |
| 5655 | godin | 150 | [1, 2, 3, 4, 5] | 0.8 | 200 | 200 | 0.001 | 0.999 | 0.7279 | 0.6866 |
| 88 | godin | 50 | [1, 2, 3, 4, 5] | 0.5 | 200 | 300 | 0.0001 | 0.9 | 0.7277 | 0.6905 |
| 2740 | godin | 100 | [1, 2, 3, 4, 5] | 0.4 | 400 | 200 | 0.0001 | 0.9 | 0.7276 | 0.6848 |
| 4360 | shin | 100 | [2, 3, 3, 3, 4] | 0.6 | 100 | 300 | 0.0001 | 0.9 | 0.7270 | 0.6909 |
| 1629 | godin | 50 | [2, 3, 3, 3, 4] | 0.5 | 200 | 400 | 0.0001 | 0.999 | 0.7270 | 0.6908 |
| 4378 | godin | 100 | [2, 3, 3, 3, 4] | 0.6 | 200 | 300 | 0.001 | 0.9 | 0.7267 | 0.6898 |
| 1341 | godin | 50 | [1, 2, 2, 2, 3] | 0.6 | 400 | 400 | 0.0001 | 0.999 | 0.7267 | 0.6862 |
| 168 | godin | 50 | [1, 2, 3, 4, 5] | 0.6 | 300 | 300 | 0.0001 | 0.9 | 0.7266 | 0.6890 |
| 2733 | shin | 100 | [1, 2, 3, 4, 5] | 0.4 | 300 | 400 | 0.0001 | 0.999 | 0.7265 | 0.6962 |
| 2876 | godin | 100 | [1, 2, 3, 4, 5] | 0.6 | 400 | 400 | 0.0001 | 0.9 | 0.7265 | 0.6863 |
| 4333 | godin | 100 | [2, 3, 3, 3, 4] | 0.5 | 300 | 400 | 0.0001 | 0.999 | 0.7264 | 0.6828 |
| 2844 | godin | 100 | [1, 2, 3, 4, 5] | 0.6 | 200 | 400 | 0.0001 | 0.9 | 0.7263 | 0.6877 |
| 1715 | godin | 50 | [2, 3, 3, 3, 4] | 0.6 | 400 | 100 | 0.001 | 0.999 | 0.7258 | 0.6845 |
| 3500 | godin | 100 | [3, 4, 5, 6, 7] | 0.4 | 300 | 400 | 0.0001 | 0.9 | 0.7257 | 0.6901 |
| 3293 | godin | 100 | [2, 3, 4, 5, 6] | 0.7 | 200 | 400 | 0.0001 | 0.999 | 0.7255 | 0.6887 |
| 7241 | godin | 150 | [2, 3, 3, 3, 4] | 0.9 | 100 | 300 | 0.0001 | 0.999 | 0.7255 | 0.6828 |
| 4247 | godin | 100 | [2, 3, 3, 3, 4] | 0.4 | 200 | 200 | 0.001 | 0.999 | 0.7254 | 0.6842 |
| 6095 | godin | 150 | [2, 3, 4, 5, 6] | 0.9 | 100 | 400 | 0.001 | 0.999 | 0.7254 | 0.6867 |
| 235 | godin | 50 | [1, 2, 3, 4, 5] | 0.7 | 300 | 300 | 0.001 | 0.999 | 0.7253 | 0.6846 |

Table 6: Best parameters for random search over shallow CNN ensembles, sorted by their performance on training data after 5-fold cross validation.

We also compare our best performing model with the top models of SMM4H shared task in Table 7. Table 5, shows the randomly chosen values of hyperparameters for the best models of each architecture. Detailed hyperparameter list of the *stacked ensemble of shallow CNN ensembles* is provided in Table 4.4.

| Systems | $(\mathcal{P}_{1+2})$ | $(\mathcal{R}_{1+2})$ | $(\mathcal{F}_{1+2})$ |
|---|---|---|---|
| Our Best Model | **0.725** | 0.664 | **0.693** |
| UKNLP | 0.701 | **0.677** | 0.689 |
| NRC-Canada | 0.704 | 0.635 | 0.668 |
| TurkuNLP | 0.701 | 0.630 | 0.663 |
| CSaRUS-CNN | 0.709 | 0.604 | 0.652 |

Table 7: Performance comparison of our system with the other state-of-the-art systems.

## 5 Conclusion and Future Work

In this paper we showed the generic effectiveness of BLSTMs and ensembles on identification of personal medication intake from Twitter posts. Our proposed architecture of stacked ensemble of shallow CNN ensembles, out-performed other models. This provided an empirical evaluation of our initial aim of combining ensembles with CNNs along with training the models using random search on the hyperparameters. In the future, we plan to work more on hyperparameter tuning using random search and various other search procedures and analyze their effectiveness. Instead of using pre-trained word embeddings it would also be interesting to look at the performance of our models by training word embeddings on a domain specific dataset of tweets. We would also like to use the classifier for studying moods and emotions of social media users expressing intake of medicine and plan to use our system in solving some of the problems that lies at the intersection of pharmacovigilance, affective computing and psychology.